\documentclass[lettersize,journal]{IEEEtran}
\usepackage{amsmath,amsfonts}
\usepackage{array}
\usepackage[caption=false,font=normalsize,labelfont=sf,textfont=sf]{subfig}
\usepackage{textcomp}
\usepackage{stfloats}
\usepackage{url}
\usepackage{verbatim}
\usepackage{cite}
\usepackage{latexsym}

\usepackage{amsmath,amssymb,amsfonts}
\usepackage{graphicx}
\usepackage{algorithm}
\usepackage{algpseudocode}
\usepackage{multirow}
\usepackage{tabularx}
\usepackage{pifont}
\usepackage{color}
\usepackage[table]{xcolor}
\usepackage{bm}

\usepackage{url}
\usepackage{xcolor}
\usepackage{dsfont}
\usepackage{marvosym}
\definecolor{newcolor}{rgb}{.8,.349,.1}

\def\citep{\cite}
\def\highlight{black}

\begin{document}

\title{SAME++: Self-supervised Anatomical eMbedding Enhanced medical image registration}
\title{SAME++: A medical image registration framework enhanced via self-supervised anatomical embeddings}

\author{Lin~Tian, Zi~Li, Fengze~Liu, Xiaoyu~Bai, Jia~Ge, Le~Lu~\IEEEmembership{Fellow,~IEEE}, Marc~Niethammer, Xianghua~Ye, Ke~Yan, and Dakai~Jin~\IEEEmembership{Member,~IEEE}
\thanks{
L. Tian and M. Niethammer are with the University of North Carolina at Chapel Hill. This work was done during L. Tian's internship at Alibaba Group. Z. Li, X. Bai, L. Lu, K. Yan and D. Jin are with  Alibaba Group. F. Liu is with Johns Hopkins University. J. Ge and X. Ye are with The First Affiliated Hospital of Zhejiang University.
}
\thanks{L. Tian and Z. Li have contributed equally to this work.}
\thanks{Correspondence: K. Yan and D. Jin (email: yankethu@gmail.com and dakai.jin@gmail.com)}
}

\markboth{Journal of \LaTeX\ Class Files,~Vol.~14, No.~8, Jan~2024}%
{Shell \MakeLowercase{\textit{et al.}}: A Sample Article Using IEEEtran.cls for IEEE Journals}


\maketitle

\begin{abstract}
Image registration is a fundamental medical image analysis task. Ideally, registration should focus on aligning semantically corresponding voxels, i.e., the same anatomical locations. However, existing methods often optimize similarity measures computed directly on intensities or on hand-crafted features, which lack anatomical semantic information. These similarity measures may lead to sub-optimal solutions where large deformations, complex anatomical differences, or cross-modality imagery exist. In this work, we introduce a fast and accurate method for unsupervised 3D medical image registration building on top of a Self-supervised Anatomical eMbedding (SAM) algorithm, which is capable of computing dense anatomical correspondences between two images at the voxel level.
We name our approach SAM-Enhanced registration (SAME++), which decomposes image registration into four steps: affine transformation, coarse deformation, deep non-parametric transformation, and instance optimization. Using SAM embeddings, we enhance these steps by finding more coherent correspondence and providing features with better semantic guidance.  SAME++ is extensively evaluated using more than 50 labeled organs on three challenging inter-subject registration tasks of different body parts (head \& neck, chest, and abdomen). Quantitative results show that SAM-affine significantly outperforms the widely-used affine registration methods by Dice score improvement of at least $4.4\%$, $6.0\%$, and $8.5\%$ for the three inter-patient registration tasks, respectively. For the non-parametric transformation step alone, SAM-deform achieves the overall best performance compared with top-ranked optimization-based and learning-based registration methods. As a complete registration framework, SAME++ markedly outperforms leading methods by $4.2\%$ - $8.2\%$ in terms of Dice score while being orders of magnitude faster than numerical optimization-based methods. Code is available at \url{https://github.com/alibaba-damo-academy/same}.
\end{abstract}


\section{Introduction}
Registration is a fundamental step in many medical image applications~\citep{ViergeverMKMSP16}, including atlas-based segmentation~\citep{asman2013non,rohlfing2004evaluation}, longitudinal lesion quantification~\citep{BuergerSK11}, image-guided radiotherapy~\citep{heinrich2013towards,jin2021deeptarget}, and computer-aided diagnosis with multi-modality fusion~\citep{LiuLFZHL22}. Its goal is to find the spatial correspondence between pairs or series of medical images. Based on the complexity of the deformation, spatial correspondences can be represented by a low dimensional parametric transformation~\citep{BuergerSK11} (e.g., for rigid or affine registration), a high-dimensional parametric/non-parametric transformation with many degrees of freedom~\citep{SotirasDP13} (e.g., represented by a spline, a displacement field, or a velocity field), or a composition of the two. 

There are mainly two kinds of registration methods. One formulates image registration as an optimization problem~\citep{Ashburner07,HeinrichJBS12,HeinrichPSH14,SunNK14,SiebertHH21}, aiming at finding the optimal transformation parameters that minimize the dissimilarity between the warped image and the fixed image, subject to certain regularity constraints. The associated optimization problems are generally non-convex and lack a closed-form solution. Thus, iterative optimization algorithms (typically based on a form of gradient descent) are commonly adopted. To reduce the computation time, learning-based methods~\citep{yang2017quicksilver,ShenHXN19,DalcaBGS19,BalakrishnanZSG19, MokC20,CVPRMokC20,LiuLFZHL22,greer2021icon,tian2022gradicon} have recently been proposed. The training of these learning-based methods also relies on optimization guided by the similarity measure and typically a regularizer. For both optimization-based or learning based methods, the similarity measures are computed either on image intensities directly or on hand-crafted features (e.g. SSC~\citep{heinrich2013towards}, MIND~\citep{HeinrichJBMGBS12} and attribute vectors~\citep{shen2002hammer}) that are designed to incorporate structural information. Another body of work~\citep{ehrhardt2010automatic,brox2010large,han2010feature} involves finding corresponding key-points in two images based on a similarity measure computed over local features (e.g., surface norm~\citep{ehrhardt2010automatic}, curvature characteristics~\citep{ehrhardt2010automatic} and histogram of oriented gradients~\citep{horn1981determining}), followed by estimating the deformation directly from the corresponding key-points.  


Both types of methods rely on the similarity measures computed either directly on intensity values or using hand-crafted features. Such measures may have limited capacity to accurately capture anatomical semantic similarity due to a lack of anatomical information in these measures. As a result, either optimizing non-convex registration energies or searching for corresponding key-points may lead to sub-optimal solutions where two voxels containing similar local structures but belonging to different anatomical regions are mismatched due to lack of anatomical semantic information.  
This could lead to severe issues when solving large deformation registration problems, where usually an affine transformation is estimated first and followed by a more flexible registration allowing for local deformations. In this case, a sub-optimal solution of the affine transformation estimated based on local features may not be able to provide a suitable initialization for the subsequent registration and may negatively affect the overall registration accuracy. The lack of anatomical information can be solved by extracting features from an anatomical segmentation (label) map~\citep{shen2002hammer}. But annotating segmentation maps is not a trivial task.

To overcome the discussed issues, we explore incorporating the Self-supervised Anatomical eMbedding (SAM)~\citep{yan2022sam} approach into registration methods. Instead of relying on manually designing local features, SAM learns a unique embedding for each voxel in the image via self-supervised contrastive learning and provides both an anatomical semantic representation and a local structure representation without using of segmentation maps. It is capable of directly finding matches between anatomical key points of two images. Moreover, it transforms the input image from the intensity space to a common feature space reducing possible contrast differences due to various imaging protocols or acquisition devices. The most straightforward way to incorporate SAM into registration is to densely extract SAM embeddings from both moving and fixed images. For each voxel in the moving image, we can then search for the matching point in the fixed image based on the most similar SAM embedding, and calculate the coordinate offsets for each pair of matching voxels as the displacement. However, this approach is computationally expensive. E.g., there are millions of voxels in a typical 3D computed tomography (CT) image. Moreover, potential mismatched voxel pairs may greatly affect the regularity of the transformation and registration accuracy.

We propose SAM-Enhanced registration (SAME++), a four stage registration framework that is based on the self-supervised pre-trained SAM. At each stage, we tailor the approach and utilize SAM in alignment with the specific traits of that stage, ultimately resulting in an accurate and fast registration framework.
First, we introduce SAM-affine, which involves the extraction of a set of corresponding points based on SAM. These key points are subsequently used to compute the affine transformation. Following that, we present SAM coarse deformation step. This stage is focused on estimation of a coarse deformation field given the key points found in the previous step. Notably, this step requires no additional training and serves as a favorable initialization for subsequent stages.
Next, we introduce SAM-deform, where we train a registration neural network to predict a dense transformation field. In this step, we enhance the network's capabilities by integrating SAM-based correlation features and leveraging a similarity measure within the SAM feature space. Lastly, we employ a SAM-based instance optimization module to counter the common generalization issue of learning-based registration methods caused by the small datasets in medical image registration.

We extensively evaluate SAME++ on more than 50 labeled organs in three challenging inter-subject registration tasks of different body parts (head \& neck, chest, and abdomen).
%
We compare SAME++ with two widely-used optimization-based methods (Elastix~\citep{KleinSMVP10}, DEEDS~\citep{heinrich2013mrf,heinrich2015multi}) and two registration methods (ConvexAdam~\citep{SiebertHH21} and LapIRN~\citep{MokC20}) that achieved the top rankings in a recent Learn2Reg challenge~\citep{hering2022learn2reg}.
Quantitative results show the superiority of SAME++: for affine registration, SAM-affine significantly outperforms the widely-used affine transformation techniques in terms of Dice score by at least 4.4\%, 6.0\%, and 8.5\% for three inter-patient registration tasks, respectively. 
%
SAM-deform achieves the overall best performance as compared with four top-ranked conventional and learning-based deformable methods under the same pre-alignment condition.
As a complete registration framework (from SAM-affine to SAM-deform), SAME++ markedly outperforms the leading methods in terms of Dice scores by 4.2\% - 8.2\% averaged on three registration tasks.

%
This work extends our previous preliminary work SAME~\citep{LiuYHGLYHXXYJ21}. Compared with~\citep{LiuYHGLYHXXYJ21}, substantial extensions are made in terms of methodology and comprehensive experimental evaluations: (1) we propose a stable sampling strategy based on cycle consistency to eliminate potential false correspondence matches of SAM embeddings in SAM-affine; (2) a regularization constraint in SAM-coarse is introduced to significantly reduce the folding rate while improving registration accuracy; (3) we incorporate diffeomorphic transformations, e.g., by using a stationary velocity field, in SAM-deform to guarantee desirable diffeomorphic properties; the deformation map is further finetuned by an auxiliary instance optimization module; (4) we conduct extensive experiments on three datasets (with more than 50 labeled organs) of different body parts to validate the performance and to compare to recent leading registration methods, such as ConvexAdam~\citep{SiebertHH21} and LapIRN~\citep{MokC20}. The paper is organized as follows. Sec.~\ref{sec:related_work} describes related work. Sec.~\ref{sec:background} introduces the background and Sec.~\ref{sec:methods} describes our framework. We show experimental results in Sec.~\ref{sec:experiments}. Sec.~\ref{sec:conclusions} provides conclusions.
\begin{figure*}[htp]
    \centering
    \includegraphics[width=0.9\textwidth]{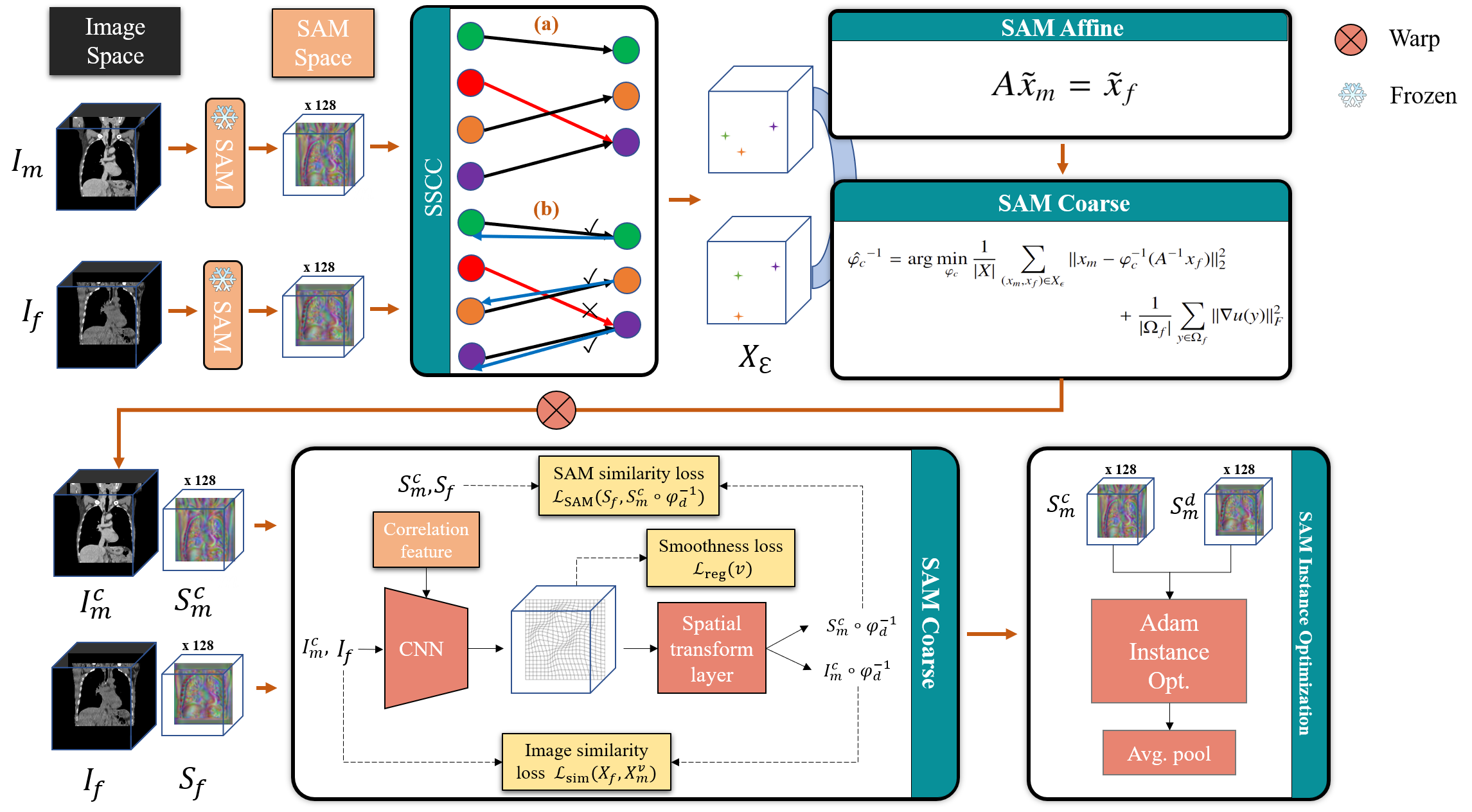}
    \caption{The framework of SAME++. Based on the SAM space, we break down image registration into four steps: keypoint-based affine transformation, coarse deformation, deep deformable registration, and instance optimization. (a) Illustration of one incorrect match based on SAM. (b) Eliminating false correspondence via cycle consistency matching.}
    \label{fig:pipeline}
     \vspace{-1.0em}
\end{figure*}
\section{Related Work} \label{sec:related_work}

\subsection{Medical Image features}
Several hand-crafted descriptors~\citep{heinrich2013towards,HeinrichJBMGBS12,shen2002hammer} have been proposed for medical image registration. SSC~\citep{heinrich2013towards} and MIND~\citep{HeinrichJBMGBS12} aim at countering intensity differences between different modalities. In~\citep{shen2002hammer}, an attribute vector containing the local edge types, image intensity, and geometric moment invariants is proposed to represent the geometric structure of the underlying anatomy. Long et al.~\citep{long2014convnets} explore the intermediate features from a classification convolutional neural network (CNN) and found that the learned features can be used to find correspondence between object instances and performs on par with hand-crafted features. Further, more work~\citep{choy2016universal,han2015matchnet,zagoruyko2015learning} has been conducted to learn dense features in a supervised training scheme, demonstrating improvements of learned CNN features over hand-crafted features. This trend has been further pursued via self-supervised visual feature learning~\citep{jing2020self}, where a pre-text task is specifically designed to learn the image representations with unlabelled data. Among which, VADeR~\citep{o2020unsupervised} and DenseCL~\citep{wang2021dense} have been developed to learn dense representations of natural images that can be used to discriminate instances across categories (e.g., dogs vs cats). Different from natural images, in medical images one would generally like to discriminate the anatomical structures instead of the subjects (patients). To achieve this goal, SAM~\citep{yan2022sam} proposes to learn a dense representation for medical images that contains \emph{anatomical semantic information} and shows the discriminativeness of the learned features by matching corresponding landmarks on several anatomical structures (e.g., chest, hand and pelvic).

\subsection{Affine Registration}
Affine registration has been extensively studied in medical image registration~\citep{avants2009advanced,jenkinson2001global,butz2001affine,ShenHXN19,HuangYLLZWZW21,ZhaoLLCX20,yu2022keymorph,MokC22}, natural image matching~\citep{lowe1999object} and point set registration~\citep{feldmar1994rigid, du2008affine,besl1992method,zhang1994iterative}. In medical image registration, the problem is generally solved by formulating an optimization problem with the affine transformation being the parameters and an intensity-based similarity being the cost function. In the point set domain, iterative closest point (ICP)~\citep{besl1992method,zhang1994iterative} iterates over the following three steps: (1) finding a set of matched closest point pairs according to the Euclidean distance, (2) estimating the affine registration parameters via least square fitting, and (3) updating the source point sets. For natural images, hand-crafted local features~\citep{lowe1999object,karami2017image} have been used to find matched point pairs. These approaches rely on either local image intensities, local features or metrics that do not take the anatomical semantic information into consideration. Recently, a number of learning-based affine registration methods have been proposed~\citep{ShenHXN19,HuangYLLZWZW21,ZhaoLLCX20,yu2022keymorph,MokC22}.
They show on par or better performance than conventional unsupervised affine registration approaches. However, they require training for each anatomical region or modality and show weaker generalizability than conventional method~\citep{MokC22}.



\subsection{Learning-based Registration}
Traditional deformable registration methods~\citep{Ashburner07,SiebertHH21,HeinrichJBS12} solve an optimization problem and iteratively minimize a similarity measure, often jointly with a regularizer, to align a pair of images.  
Recently, learning-based deformable registration~\citep{yang2017quicksilver,BalakrishnanZSG19, MokC20,CVPRMokC20,LiuLFZHL22,ShenHXN19,HuangYLLZWZW21,ZhaoLLCX20}, using deep networks, has been investigated. Compared with optimization-based approaches learning-based methods are much faster at inference.
Quicksilver~\citep{yang2017quicksilver} and Voxelmorph~\citep{BalakrishnanZSG19} were the initial approaches for supervised and unsupervised medical image registration, where a convolutional neural network predicts a vector-field to directly describe the displacements or to obtain a velocity field from which a transformation can be obtained by integration. To be able to capture large deformations, more recent methods~\citep{ShenHXN19,CVPRMokC20,LiuLFZHL22,MokC20,HuangYLLZWZW21,ZhaoLLCX20,LiuLZFL20,10231109} focus on designing sophisticated networks using multi-step approaches, pyramids, cascaded structures or connecting registration to image synthesis or segmentation~\citep{xu2019deepatlas}. Compared to the extensive work on network structures, less work explores the similarity measure. Registration performance can be improved by using the segmentation (label) maps within the loss~\citep{BalakrishnanZSG19,xu2019deepatlas}. Compared to similarity measures computed over image intensities, segmentation maps can provide anatomical information during training. However, anatomical segmentation maps are not always available. Different from using explicit anatomical segmentation maps, our work explores using pre-trained SAM features, that contain discriminative anatomical semantic information.

\section{Background}\label{sec:background}

\subsection{Problem formulation}
Given two images $I_m:\Omega_m\to{\mathbb{R}}$ and $I_f:\Omega_f\to{\mathbb{R}}$, $\Omega_m\subset{\mathbb{R}^n}$ and $\Omega_f\subset{\mathbb{R}^n}$ representing the domains of the moving and fixed images respectively, our goal is to find the spatial transformation $\varphi:\Omega_m\to\Omega_f$ which makes the warped moving image $I_m\circ\varphi^{-1}$ as similar as possible to the fixed image $I_f$. 
Note that the dimension $n$ of the domains $\Omega\subset{\mathbb{R}^n}$ can be 2D or 3D. In the following sections, we assume $n=3$ without loss of generality.

\subsection{Self-supervised anatomical embedding (SAM) review}
SAM~\citep{yan2022sam} is a voxel-wise contrastive learning framework to encode the semantic anatomical information of
each voxel, so that the same anatomical location across different images will have similar embeddings. With a coarse-to-fine network structure and a hard-and-diverse negative sampling strategy, SAM learns one global and one local feature embedding per voxel in a given image. Each feature embedding expresses the deterministic semantic representation per voxel. The learned SAM feature has demonstrated efficacy in various downstream tasks, e.g., for anatomical point matching, landmark detection, and longitudinal lesion matching~\citep{cai2021deep,hering2021whole}. 

Because of the semantic meaning of SAM features, they can directly be used in registration. For any image $I$ with shape $D\times{H}\times{W}$, SAM extracts a global feature map and a local feature map with size $C\times\frac{D}{2}\times\frac{H}{8}\times\frac{W}{8}$ and $C\times\frac{D}{2}\times\frac{H}{2}\times\frac{W}{2}$ respectively with $C$ being the dimension of the feature embedding at each voxel. In our work, we adopt the pre-trained SAM from ~\citep{yan2022sam}, resize the global feature map to the same size $C\times\frac{D}{2}\times\frac{H}{2}\times\frac{W}{2}$ as the local feature map by linear interpolation and normalize the feature embedding via L2 normalization. Then we concatenate the resized global feature map with the local feature map along the channel dimension, resulting in the final SAM feature map that is used in our work. We denote the SAM feature maps of $I_m$ and $I_f$ as $S_m$ and $S_f$, respectively,
and define $d(\cdot,\cdot)$ to measure the similarity between two SAM feature embeddings. Since feature embeddings are normalized before concatenation, we use the dot product as the measure $d(\cdot,\cdot)$, which corresponds to cosine similarity. A higher score indicates that the pixels at corresponding locations are anatomically more similar. To be noted, the pre-trained SAM is kept frozen.
\section{Methods}\label{sec:methods}
As shown in Fig.~\ref{fig:pipeline}, SAME++ consists of four consecutive steps. The initial step involves SAM-affine and SAM-coarse, which aim to seek the optimal affine transformation and a coarse displacement field that can match a set of corresponding points between $I_m$ and $I_f$ extracted based on a measure $d(\cdot, \cdot)$ in SAM space.
Following SAM-affine and SAM-coarse, SAM-deform leverages a neural network to predict a transformation field, which is then optimized using an auxiliary instance optimization module.

\subsection{SAM-Affine}\label{sec:affine}
SAM features allow us to extract corresponding points between $I_m$ and $I_f$, and to estimate the affine transformation matrix given the corresponding points. To achieve this, we start by selecting an initial set of points 
$\{x_m|x_m\in\Omega_m, m=0,1...N\}$ 
\textcolor{\highlight}{, which are evenly distributed} in the domain of the moving image. The most straightforward approach of finding the corresponding points is to search for points $x_f\in\Omega_f$ that have the most similar SAM embeddings to the points $x_m$. However, mismatched pairs of points could exist due to inaccurate SAM embeddings. 
In our preliminary work~\citep{LiuLFZHL22}, we address this issue via thresholding the similarity $d(\cdot,\cdot)$ between the corresponding points.

In this work, we further reduce the incorrect correspondences via cycle consistency (SSCC) that computes a set of stable SAM matched points from $\Omega_m$ and $\Omega_f$. Specifically, for a point $x_m\in\Omega_m$, we first find the matching point $x_f\in\Omega_f$ via \Call{findpoints}{$\{x_m\}, S_m, S_f$} (Alg.~\ref{Alg:SSCC}). Then we compute the corresponding point $x_m^{\prime}\in\Omega_m$ to $x_f$ via \Call{findpoints}{$\{x_f\}, S_f, S_m$}. Presumably, if $(x_m,x_f)$ is a correct matching pair, then $x_m$ and $x_m^{\prime}$ should be the same point. Otherwise, there is a high chance that $(x_m,x_f)$ is not a corresponding pair. This idea is illustrated in Fig.~\ref{fig:pipeline}. 
To rule out the mismatched pairs, we substitute $x_m$ with $x_m^{\prime}$ and repeat the process for $K$ times. This stable sampling algorithm is outlined in Alg.~\ref{Alg:SSCC}. In Alg.~\ref{Alg:SSCC}, \Call{selectpoints}{$\Omega$} is a function to select a set of points in the given image domain. And \Call{findpoints}{$\{x_k\}, S_k, S_q$} searches on the grids of the query feature maps $S_q$ for the point that has the most similar SAM embedding to the feature vector of the key point $S_k(X_k)$. In practice, line~\ref{algo:sscc_start} to line~\ref{algo:sscc_end} in Alg.~\ref{Alg:SSCC} are computed in parallel via a convolution operation.
 
After determining the corresponding points $X=\{(x_m, x_f)|x_m\in\Omega_m, x_f\in\Omega_f\}$, we further remove the low-confidence matched pairs by filtering $X$ with a similarity threshold $\epsilon$, resulting in the final set $X_{\epsilon}=\{(x_m, x_f)|d(s_m(x_m),s_f(x_f))>\epsilon, (x_m,x_f)\in{X}\}$. With $X_{\epsilon}$, one can solve the following linear system to obtain the affine transformation matrix:
\begin{equation}
     A\tilde{x}_m = \tilde{x}_f,
\end{equation}
where $\tilde{x}_m$ and $\tilde{x}_f$ are the homogeneous representations of $x_m$ and $x_f$, respectively, and $A\in\mathds{R}^{4\times{4}}$ is the affine transformation matrix.



\begin{algorithm}[!t]
    \caption{Stable Sampling via Cycle Consistency}
    \label{Alg:SSCC}
    \begin{algorithmic}[1]
        \Require  
		$S_m, S_f$: SAM features of the moving and fixed image;
            $K$: The number of iterations.
	\Ensure 
		$\{x_m\}$: A list of points in $\Omega_m$;
		$\{x_f\}$: A list of matched points in $\Omega_f$.
        
        \Function{FindPoints}{$\{x_k\}$, $S_k$, $S_q$}
        \State Initialize $X_q=\{\}$
        \For {$x_k \in \{x_k\}$} \label{algo:sscc_start}
            \State Initialize $D=\{\}$
            \For {Each point $x_q$ on the grids of $S_q$}
            \State Add $d(S_k(x_k),S_q(x_q))$ into $D$
            \EndFor
            \State Add the $x_q$ corresponding to \Call{Max}{$D$} to $X_q$
        \EndFor \label{algo:sscc_end}
        \State \Return $X_q$
        \EndFunction

        \State $\{x_m\} =$ \Call{selectpoints}{$\Omega_m$}
        \For{$k \gets 1$ to $K$}
            \State $\{x_f\}$ $\gets$  \Call{FindPoints}{$\{x_m\}$, $S_m$, $S_f$}
            \State $\{x_m\}$ $\gets$  \Call{FindPoints}{$\{x_f\}$, $S_f$, $S_m$}
        \EndFor
        \State \Return $\{x_m\}, \{x_f\}$.
    \end{algorithmic}
\end{algorithm}
\vspace{-1.0em}
\subsection{SAM-coarse}
After SAM-affine, we introduce an additional SAM-coarse step to deform the moving image based on the corresponding point pairs $X$\textcolor{\highlight}{$_{\epsilon}$}. Compared to an affine pre-alignment, SAM-coarse provides local warps that can serve as a better initialization for a following learning-based registration.
SAM-coarse is implemented in a similar setting as SAM-affine but solves for a coarse displacement field (DVF). Given the set of SAM matched point pairs $X_{\epsilon}$ computed in Sec.~\ref{sec:affine}, we optimize over a coarse DVF $\varphi_c^{-1}=x+u(x), \varphi_c^{-1}:\Omega_f\to\Omega_m$, which aims to bring the SAM matched points $A^{-1}x_f$ and $x_m$ closer. To be noted, $\varphi_c^{-1}$ is a coarse DVF with shape $3\times\frac{D}{stride}\times\frac{H}{stride}\times\frac{W}{stride}$. We set $stride=4$ in the experiments. This step accounts for deformations that cannot be explained by an affine transformation matrix. 

In our preliminary work~\citep{LiuLFZHL22}, we proposed to directly estimate the coarse DVF from the displacement between the corresponding points. However, we found that this might introduce severe irregular deformations.
This is because there may be mismatched pairs remaining in the corresponding point set, especially when the structure at a point is not deterministic, despite our proposed SSCC approach to remove such pairs. Hence, to address this issue,
we improve SAM-coarse via formulating it as an optimization problem and use a regularizer to obtain the coarse DVF. With such a design, SAM-coarse is formulated as
\begin{multline}
    \hat{\varphi_c}^{-1} =  \arg\min_{\varphi_c}\frac{1}{|X|}\sum_{(x_m,x_f)\in{X_{\epsilon}}}||x_m-\varphi_c^{-1}(A^{-1}x_f)||^2_2 \\
    + \frac{1}{|\Omega_f|}\sum_{y\in\Omega_f}||\nabla{u(y)}||_F^2\,.
\end{multline} 
\subsection{SAM-deform}
SAM-affine and SAM-coarse estimate an affine transformation and a coarse displacement field, respectively. To improve the registration accuracy, SAM-deform is developed that aims at estimating a dense non-parametric transformation map $\varphi_d^{-1}(x)$ given the fixed image $I_f$ and a pre-aligned moving image $I_m^c$ after SAM-affine and SAM-coarse. \textcolor{\highlight}{Deep neural networks often use pure pixel intensity-based features and similarity losses to learn $\varphi_d^{-1}(x)$, such as normalized cross-correlation (NCC).  However, the NCC loss only compares local image intensities, which may not be robust under CT contrast injection, pathological changes, and large or complex deformations in the two images. On the other hand, the SAM embeddings can uncover semantic similarities between two pixels. Hence, we improve them by leveraging the semantic information contained in SAM embeddings using SAM correlation features and a SAM loss.} 

Specifically, we train a neural network $f_\theta(I_f, I_m^c, S_f, S_m^c)$ to predict the dense transformation map, as illustrated in Fig.~\ref{fig:pipeline}. To train such a network, we propose a loss containing three terms: a similarity term $\mathcal{L}_{sim}$ that penalizes the appearance differences, a similarity term $\mathcal{L}_{SAM}$ in SAM space, and a regularizer $\mathcal{L}_{reg}$ that encourages spatial smoothness of the chosen transformation model. SAM-deform is independent on the choice of similarity measure used in $\mathcal{L}_{sim}$. In the experiment, we test two commonly used similarity measure: NCC and local normalized cross-correlation (LNCC~\citep{BalakrishnanZSG19}). $\mathcal{L}_{SAM}$ is computed over the SAM features, which can be written as:
\begin{equation}
\mathcal{L}_{SAM} =\sum{1-d(S_m^c\circ\varphi_d^{-1}, S_f)}\,.
\end{equation}

Penalizing the dissimilarity between the fixed image and the warped moving image could yield a perfect match but with physically unfeasible deformations. Based on the transformation model adopted in the network, the regularizer is therefore required to obtain smooth transformations. In our preliminary work~\citep{LiuLFZHL22}, we use DVF as the transformation model. To further advocate a regular transformation. In this work, we adopt the stationary velocity field (SVF)  



\begin{equation}\label{euq:svf}
\frac{\partial \bm{\phi}(t)}{\partial t} = \bm{v}(\bm{\phi}(t)), \ \bm{\phi}(0) = Id, \bm{\varphi}=\bm{\phi}(1)\,
\end{equation}
as the transformation.
Govern by the ordinary differential equation constraint, $\bm{\phi}(0) = Id$ is the identity transformation and $t \in [0,1]$ represents the integration time. We follow~\citep{dalca2018unsupervised} to obtain the final registration field $\bm{\varphi}$ via integration using scaling and squaring~\citep{Ashburner07}, which recursively computes the solution in successive small time steps. We use seven steps in our experiments. We adopt the following regularization to advocate a smooth velocity field
\begin{equation}
    \mathcal{L}_{reg}=\sum\frac{1}{|\Omega|}\sum_{x\in\Omega}||\nabla{\bm{v}(x)}||_F^2.
\end{equation}
The overall loss is 
\begin{equation}
    \mathcal{L} = \lambda_1(\mathcal{L}_{sim} + \mathcal{L}_{SAM}) + \lambda_2\mathcal{L}_{reg}
\end{equation}
where $\lambda_1$ and $\lambda_2$ is the hyper-parameters to balance the loss terms.
\subsection{SAM Instance Optimization}
Upon our preliminary work~\citep{LiuLFZHL22}, we further improve the registration performance by adding instance optimization module, which solves a conventional optimization problem
\begin{equation}
    \hat{\varphi_i} =  \arg\min_{\varphi_i}1-d(S_m^c\circ\varphi_i^{-1},S_f)
     + \frac{1}{|\Omega|}\sum_{\Omega}||\nabla{u_i(x)}||_2^2\,,
\end{equation}
where the transformation is defined as $\varphi_i^{-1}(x) = x+u(x)$, and $S_m^c$ and $d(\cdot,\cdot)$ represent the SAM feature map of the warped image $I_m^c$ and the dot product between SAM feature vectors. The resulting transformation of SAM-deform $\varphi^{-1}_d$ are used as the initial transformation in this step.

The final transformation field is computed via composition of $A$, $\varphi_c$ and $\varphi_i$ as defined 
\begin{equation}
    \varphi^{-1} = \varphi_a^{-1}\circ\varphi_c^{-1}\circ\varphi_i^{-1},\varphi_a^{-1}(x) = A^{-1}\tilde{x}\,.
\end{equation}
We use trilinear interpolation in the composition.
\begin{table*}[!t]
\centering
\caption{Performance comparison of SAM-affine and SAM-coarse in three datasets.}
\label{tab:comparsion_affine}
\begin{tabular}{m{3.6cm}<{\centering} m{1.0cm}<{\centering} m{1.0cm}<{\centering} m{1.0cm}<{\centering} m{1.0cm}<{\centering} m{1.0cm}<{\centering}m{1.0cm}<{\centering}  m{1.0cm}<{\centering} m{1.0cm}<{\centering}m{1.0cm}<{\centering}}
\hline
 & \multicolumn{3}{c}{Chest CT } & \multicolumn{3}{c}{Abdomen CT} & \multicolumn{3}{c}{Neck CT} \\ \cline{2-10} 
 & DICE$\uparrow$ & $\%|J|\downarrow$ & Time(s) & DICE$\uparrow$ & $\%|J|\downarrow$ & Time(s) & DICE$\uparrow$ & $\%|J|\downarrow$ & Time(s)\\ \hline
Initial & 12.99 & - & - & 25.88 & - & - & 11.83 & - & - \\
MIND-affine & 28.24 & - & 6.89 & 23.62 & - & 9.70 & 18.31 & - & 10.28\\
Elastix-affine & 28.68 &  - & 7.76 & 21.90 & -  & 8.73  & 25.37 & - & 10.67 \\ \hline
SAM-affine & 32.64 & - & 0.57 &29.67   & - & 1.03 & 33.91 & - & 0.66 \\ 
SAM-A + SAM-C & 45.14 & 0.08 & 2.62 & 40.14   & 1.87  & 6.30 & 48.07 & 0.25  &  2.21 \\
\hline
\end{tabular}
\end{table*}

\begin{figure*}[!t]
    \centering
    \includegraphics[width=\textwidth]{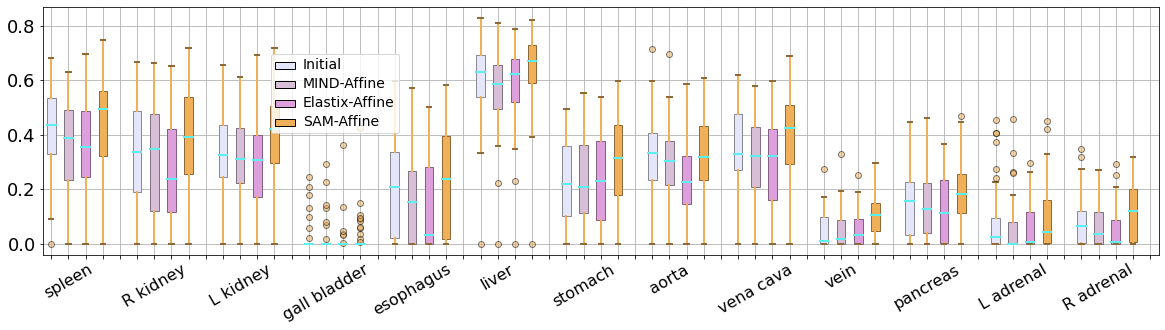}
     \vspace{-1.8em}
    \caption{Boxplot of the performance of different affine registration methods on 13 organs in the abdomen CT dataset.}
    \label{fig:abdomen_boxplot_affine}
\end{figure*}

\section{Experiments}\label{sec:experiments}
We evaluate SAME++ on three challenging inter-patient image registration tasks, each of which focuses on a specific body part.
We first describe the datasets, preprocessing, evaluation metrics, and the implementation details. Then, a series of ablation results are presented to demonstrate the effectiveness of each step of SAME++. Comparisons with the leading registration methods are also reported.

\subsection{Datasets and Applications}
\subsubsection{Head \& Neck CT}
A CT dataset containing 72 head \& neck cancer patients~\citep{Guo2021ComprehensiveAC} (denoted as Head \& Neck CT) was utilized. 13 head \& neck organs are manually labeled including the brainstem, left eye, right eye, left lens, right lens, optic chiasm, left optic nerve, right optic nerve, left parotid, right parotid, left temporomandibular joint (TMJ), right TMJ and spinal cord. We randomly split the dataset into 52, 10, and 10 for training, validation, and testing, respectively. For validation and testing, 90 image pairs are constructed for \emph{inter-patient registration}. Each image is resampled to an isotropic resolution of 2 mm and cropped to $256 \times 128 \times 224$ (mainly contains the head \& neck region) by removing black borders.

\subsubsection{Chest CT}
A chest CT dataset of 94 subjects~\citep{guo2021deepstationing} was collected, each with a contrast-enhanced and a non-contrast scan. Each chest CT image has 35 anatomical structures manually labeled (including lung, heart, airway, esophagus, aorta, bones, muscles, arteries, and veins). We randomly split the patients into 74, 10, and 10 as training, validation, and test sets. For validation and testing, 90 image pairs are constructed for \emph{inter-patient registration}, including intra-phase registration and cross-phase registration. Each image is resampled to an isotropic resolution of 2 mm and cropped to $208 \times 144 \times 192$ (mainly contains the chest region) by removing black borders.

\subsubsection{Abdomen CT}
We use the abdominal CT dataset~\citep{xu2016evaluation,hering2022learn2reg} to evaluate the \emph{inter-patient registration} of abdominal CT images.
\footnote{To be noted, we use the validation set in learn2reg~\citep{hering2022learn2reg} as our test set and split the learn2reg train set for training and validation.}{The dataset contains 30 CTs and we split it with 20 for training/validation and 10 for testing.} Each image has 13 manually labeled anatomical structures: spleen, right kidney, left kidney, gall bladder, esophagus, liver, stomach, aorta, inferior vena cava, portal and splenic vein, pancreas, left adrenal gland and right adrenal gland. The images are resampled to the same voxel resolution of 2 mm and cropped to the spatial dimensions of $192 \times 160 \times 256$ mainly containing the abdominal region.

%

\textbf{Evaluation metrics}: To evaluate the accuracy, we use the average Dice score (DICE) over the labeled organs in each dataset. 
To evaluate the plausibility of deformation fields, we compute the percentage of foldings ($\%|J|$) inside the deformation field. A negative determinant of the Jacobian at a voxel indicates local folding~\citep{Ashburner07}.
Efficiency is measured based on model inference time on the same hardware.

%
%

\subsection{Implementation Details}
We use a pre-trained SAM model~\citep{yan2022sam}.
This SAM model outputs a 128 dimensional global embedding and a 128 dimensional local embedding for each voxel and is the same in all four steps of SAME++. Image intensities in all datasets are normalized to $[-1, 1]$ using a window of $(-800,400)$ Hounsfield Unit.
In SAM-affine and SAM-coarse, the SAM similarity threshold is set to 0.7 based on the performance on ChestCT validation set and kept the same across all the datasets. In SAM-Deform, we use a 3D U-Net~\citep{CicekALBR16} as the backbone and concatenate the correlation feature and images before the convolutional block. 
We test NCC and LNCC as the similarity measure used in $\mathcal{L}_{sim}$. For experiments of Head and Chest dataset, we use NCC. LNCC is used for Abdomen dataset. The loss weights $\{\lambda_1,\lambda_2\}$ for Head, Chest and Abdomen are empirically set to $\{1,100\}$, $\{1,50\}$ and $\{0.01,10\}$ according to the performance on the validation set. 
We train SAM-Deform using the Adam optimizer with a learning rate of $5e-5$ and a batch size of 2 for about $40000$ iterations. The experiments are run on an Intel Xeon Platinum 8163 CPU with 16 CPU cores at 2.50GHz and the GPU is an NVIDIA Tesla V100.

\begin{table*}[htp]
\centering
\caption{Performance comparison of SAM-deform and other deformable registration methods. Note that inputs for all deformable methods are the SAM-coarse registration results (a better initial alignment than conventional affine transformation). mDICE denotes the mean DICE over three datasets.}
\label{tab:comparsion}
\begin{tabular}{m{3.0cm}<{\centering} m{0.9cm}<{\centering} m{0.9cm}<{\centering} m{.9cm}<{\centering} m{.9cm}<{\centering} m{0.9cm}<{\centering} m{0.9cm}<{\centering}  m{0.9cm}<{\centering} m{0.9cm}<{\centering} m{0.9cm}<{\centering} c}
\hline
 & \multicolumn{3}{c}{Chest CT } & \multicolumn{3}{c}{Abdomen CT} & \multicolumn{3}{c}{Head \& Neck CT} & \multirow{2}{*}{mDICE$\uparrow$} \\ \cline{2-10} 
 & DICE$\uparrow$ & $\%|J|\downarrow$   & Time(s)  & DICE$\uparrow$ & $\%|J|\downarrow$  & Time(s) & DICE$\uparrow$ & $\%|J|\downarrow$  & Time(s) & \\ \hline
Initial: SAM-coarse & 45.14 & - & - & 40.14   & - & - & 48.07 & - & - & 44.45\\
NiftyReg & 51.58 & 0.04 & 186.54 & 43.14 & 0.04 & 281.07 & 57.54 & 0.01 &  188.38 & 50.75 \\
Deeds  & 52.72 & 1.28 & 57.89 & 46.52 & 0.75 & 45.21 & 62.34 & 0.03 &  42.05 & 53.86\\
ConvexAdam & 54.62 & 0.73 &  6.06 & 44.44 & 2.17 &   8.83 & 61.45 & 0.31 & 5.33 & 53.50\\
LapIRN &  55.87 & 4.33  & 2.67 & 46.44  & 2.62  & 6.39 & 60.16 & 1.59 &  2.26 & 54.16\\
\hline
SAM-deform & 55.36 &  0.40 & 4.01 & 47.12 & 2.51 & 7.91 &  61.35 & 0.35 & 4.00 & \textbf{54.61}\\
\hline
\end{tabular}
\end{table*}

\begin{figure*}[htp]
    \centering
    \includegraphics[width=\textwidth]{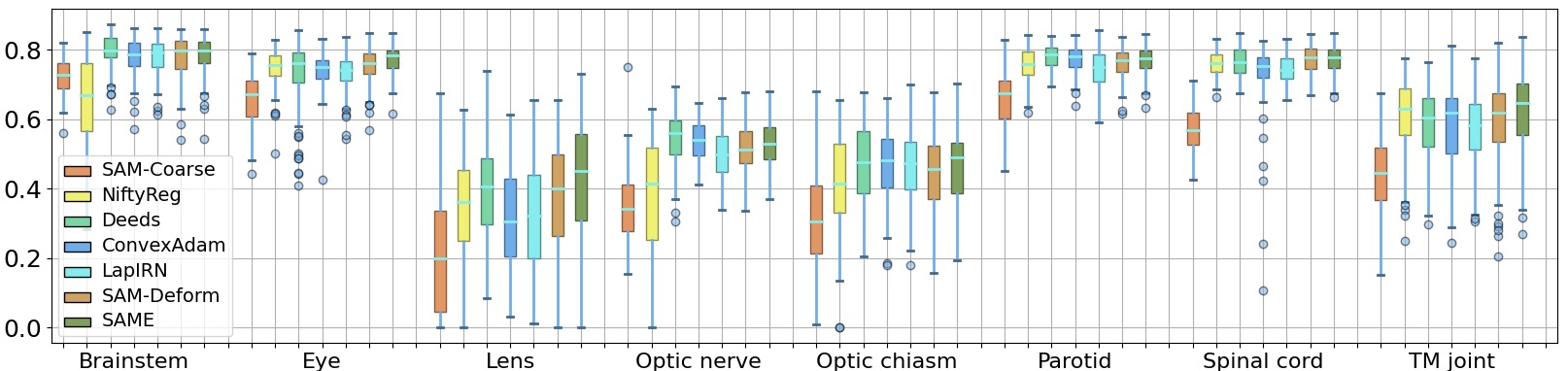}
     \vspace{-1.0em}
    \caption{Comparison of deformable registration methods on all organ groups on the neck dataset using boxplot.}
    \label{fig:neck_boxplot}
\end{figure*}

\subsection{SAM-Affine and SAM-coarse Registration Results}
We first evaluate the effectiveness of SAM-affine. Two widely-used affine registration methods are compared: (1) an intensity optimization-based method implemented in Elastix~\citep{KleinSMVP10}, and (2) a regression-based solution using MIND features~\citep{HeinrichJBMGBS12}. The quantitative results are summarized in Table.~\ref{tab:comparsion_affine}.
The comparison reveals that SAM-affine significantly outperforms the affine registration in Elastix~\citep{KleinSMVP10} with an improvement in DICE ranging from 4.4\% to 15.6\% across the three tasks. It is also markedly better than affine registration with the hand-crafted MIND~\citep{HeinrichJBMGBS12} descriptor, achieving an average improvement of 6.8\% in DICE across the three tasks. The superior of SAM-affine can be attributed to the better anatomical correspondence provided by SAM. As SAM-affine directly calculates the affine matrix by least squares fitting, it has an average inference time of 0.75s per paired image across the three datasets (more than 10 times faster compared to Elastix-affine (9.05s) or MIND-affine (8.96s)). 
%
%

To better understand the alignment of detailed anatomical structures, we plot the boxplot of 13 organs in the most challenging abdomen affine registration task in Fig.~\ref{fig:abdomen_boxplot_affine}. As shown, SAM-affine is consistently better than the two competing methods on every examined abdominal organ where large deformations and complex anatomical differences exist. We also observe that traditional affine methods may even perform worse than using no alignment (the initial condition) on several organs, e.g., spleen, pancreas, esophagus, and aorta. In contrast, SAM-affine can consistently improve the alignment. This demonstrates the importance of global semantic information (as provided by SAM) in affine registration. 

%

The effectiveness of SAM-coarse is also illustrated in Table.~\ref{tab:comparsion_affine}. After SAM-affine, SAM-coarse can further boost the registration accuracy significantly by $10.47\%$ to $14.16\%$ in DICE across different datasets.  This is because SAM-coarse allows for local deformations that provide more degrees of freedom. This step can provide a good initialization to the following step that aims to find a dense deformation map.

\subsection{SAM-deform Registration Results}
To evaluate the performance of SAM-deform, we compare with four widely-used and top-performing non-rigid registration methods. They include  NiftyReg~\citep{SunNK14}, Deeds~\citep{HeinrichJBS12}, and two top-ranked methods (ConvexAdam~\citep{SiebertHH21} and LapIRN~\citep{MokC20}) in the recent Learn2Reg~\citep{hering2022learn2reg} registration challenge. For a fair comparison, note that all deformable methods in this subsection use the pre-alignment of SAM-affine and SAM-coarse transformation, which provide better pre-alignment than other commonly adopted affine methods. 





\begin{table*}[htp]
\centering
\caption{Comparison with widely-used leading registration pipelines (from affine to deformable transformtion). The unaligned initial data is used as input to all methods.}
\label{tab:comparsion_pipline}
\begin{tabular}{m{2.2cm}<{\centering} m{0.9cm}<{\centering} m{0.9cm}<{\centering} m{.9cm}<{\centering} m{.9cm}<{\centering} m{0.9cm}<{\centering}m{0.9cm}<{\centering}  m{0.9cm}<{\centering} m{0.9cm}<{\centering}m{0.9cm}<{\centering} }
\hline
 & \multicolumn{3}{c}{Chest CT } & \multicolumn{3}{c}{Abdomen CT} & \multicolumn{3}{c}{Head \& Neck CT} \\ \cline{2-10} 
 & DICE$\uparrow$ & $\%|J|\downarrow$  & Time(s)  & DICE$\uparrow$ & $\%|J|\downarrow$ & Time(s) & DICE$\uparrow$ & $\%|J|\downarrow$ & Time(s) \\ \hline
Initial & 12.99 & - & - & 25.88   & - & - & 11.83 & - & -\\
NiftyReg & 51.65 & 0.02 & 388.28 & 33.39 & 0.09 & 478.05 & 59.92 & 0.01 &  298.10 \\
Deeds  & 52.32 & 0.82 & 276.42 & 48.31 & 1.24 & 160.32 & 56.32 & 0.25 & 185.63 \\
ConvexAdam & 52.10 & 1.27 & 13.82 & 34.42 & 1.36 & 17.56 & 58.21 & 0.30 & 16.00 \\
\hline
SAME++ &\textbf{56.93} &  0.44 & 8.74 & \textbf{49.27} & 2.82 &  11.16 &  \textbf{63.22} & 0.33 & 9.87 \\
\hline
\end{tabular}
\end{table*}

\begin{figure*}[htp]
    \centering
    \includegraphics[width=\textwidth]{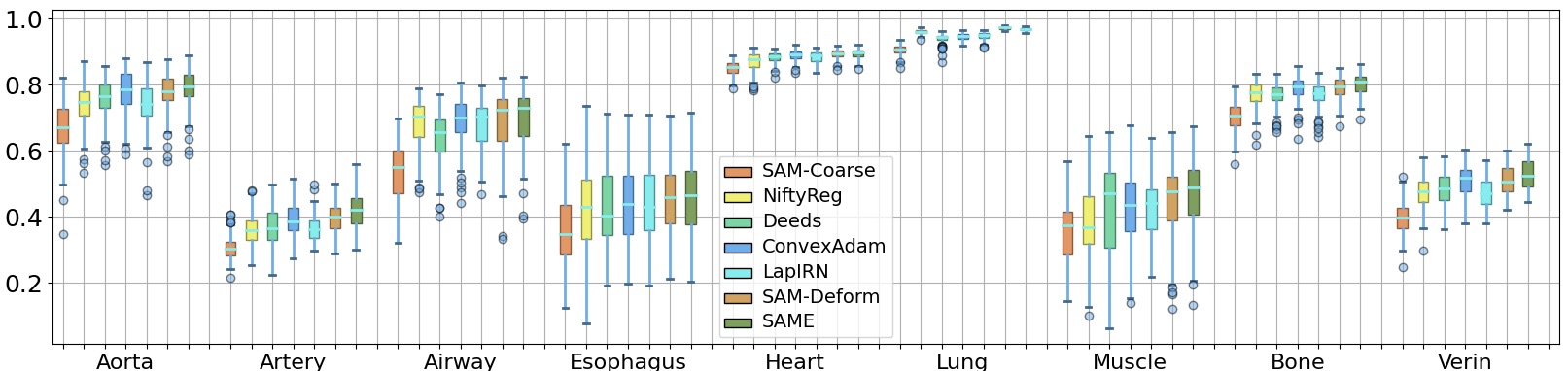}
     \vspace{-1.0em}
    \caption{Comparison of deformable registration methods on all organ groups on the chest dataset using boxplot.}
    \label{fig:chest_boxplot}
\end{figure*}

Results for the deformable registration methods are shown in Table.~\ref{tab:comparsion}. Several conclusions can be drawn. First, SAM-deform  outperforms the widely-used NiftyReg across the three datasets with 3\% average DICE improvement. Second, compared with the best traditional optimization-based method (Deeds), SAM-deform performs slightly better over the three datasets with comparable folding rate, while being ~10 times faster. Third, although the best learning-based method (LapIRN) has a comparable inference time to SAM-deform, LapIRN has a notably higher folding rate overall. Finally, SAM-deform achieves the overall best performance (54.61\% mean DICE) compared to other deformable methods (50.75\% to 54.16\% mean DICE) with the fastest inference time and a comparable folding rate.


%

\begin{figure}[!ht]
	\centering
	\begin{tabular}{c@{\extracolsep{0.12em}}c@{\extracolsep{0.12em}}c@{\extracolsep{0.12em}}c}
		
	\includegraphics[width=0.12\textwidth]{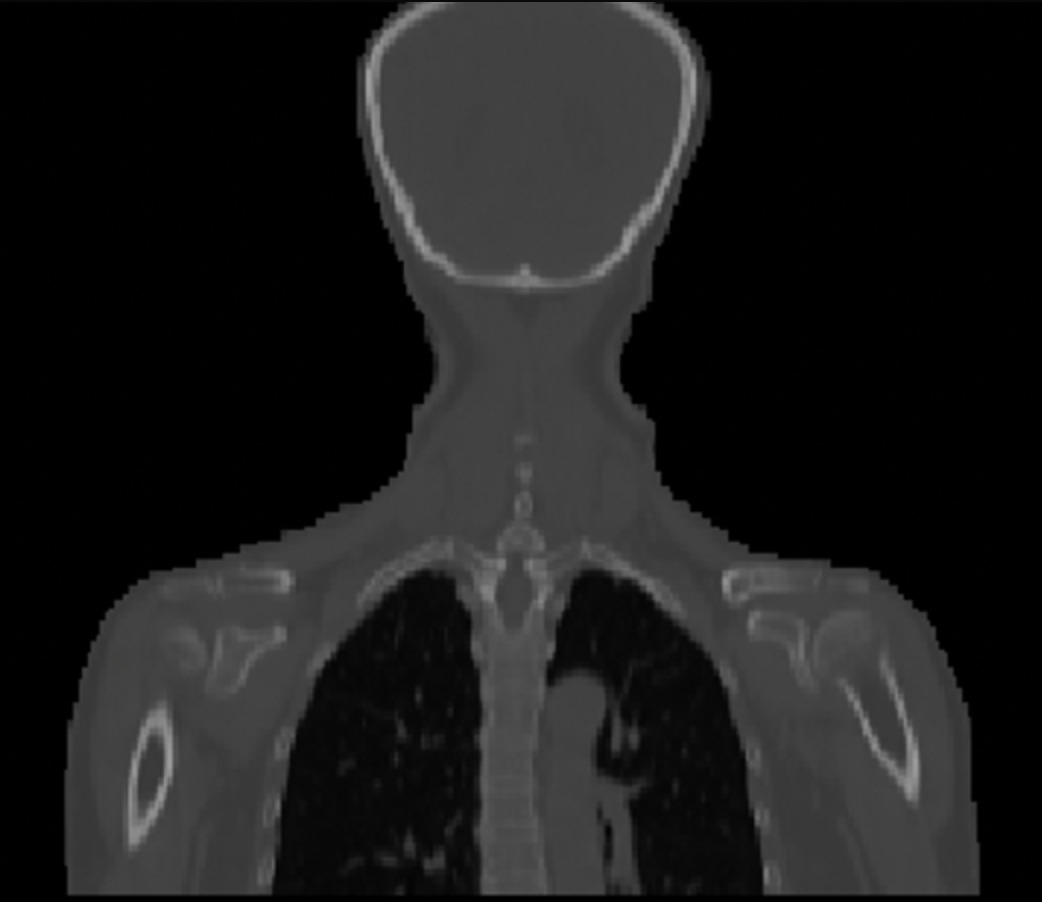}
		&\includegraphics[width=0.12\textwidth]{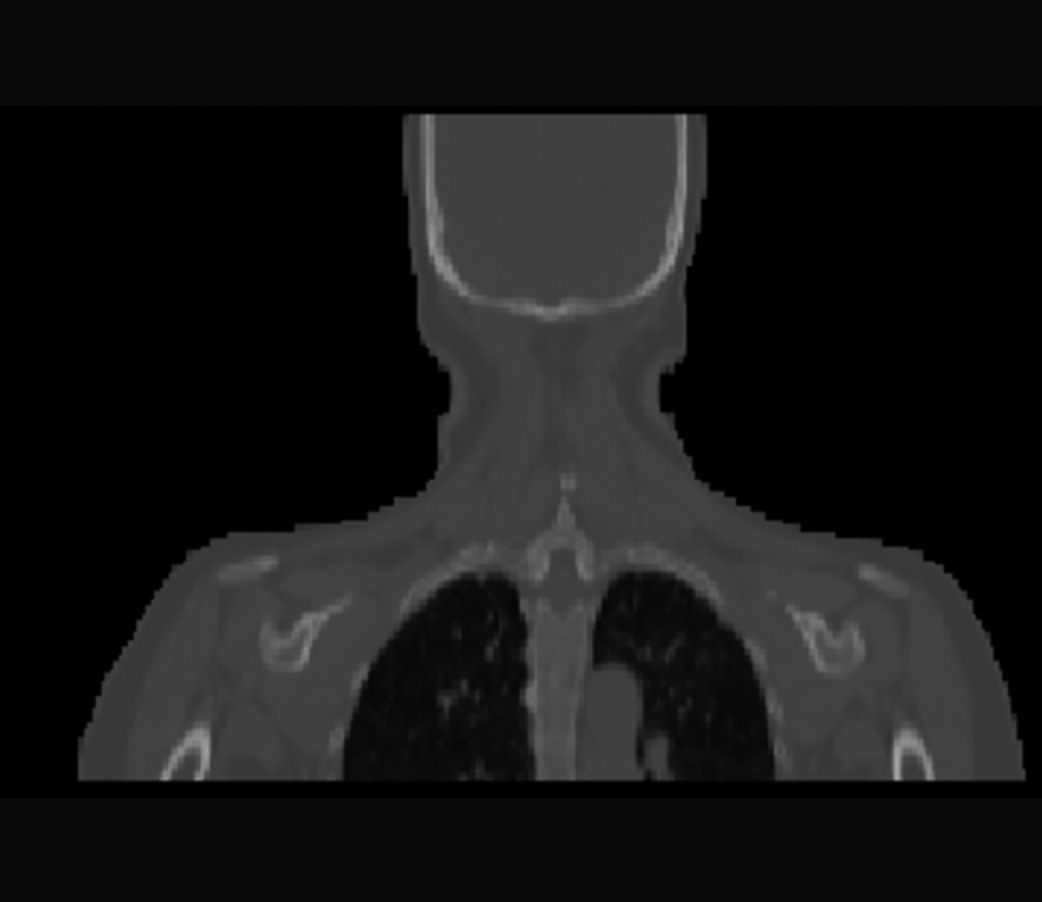}
		&\includegraphics[width=0.12\textwidth]{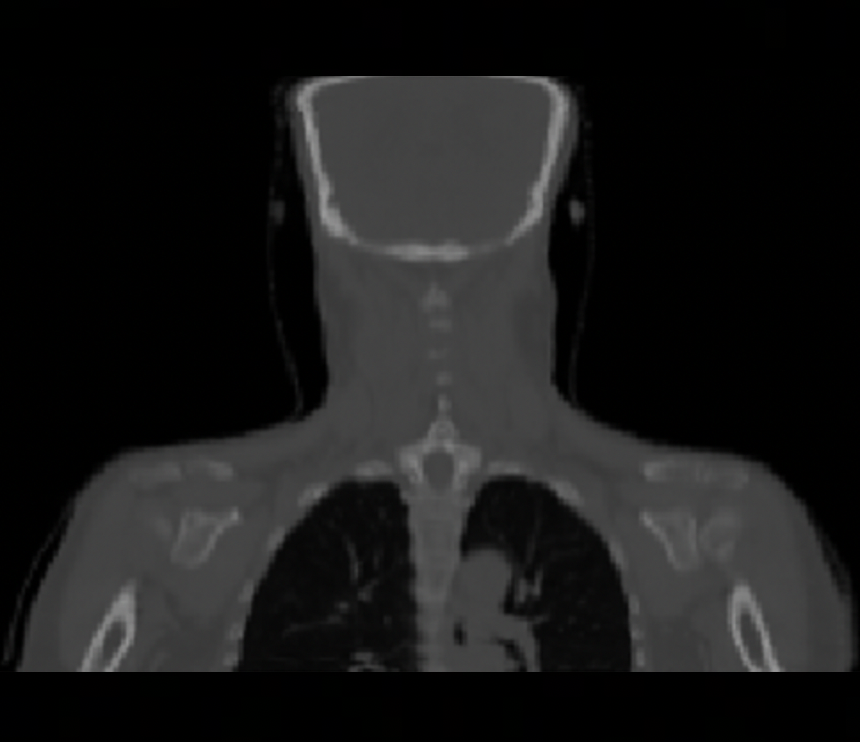}
		&\includegraphics[width=0.12\textwidth]{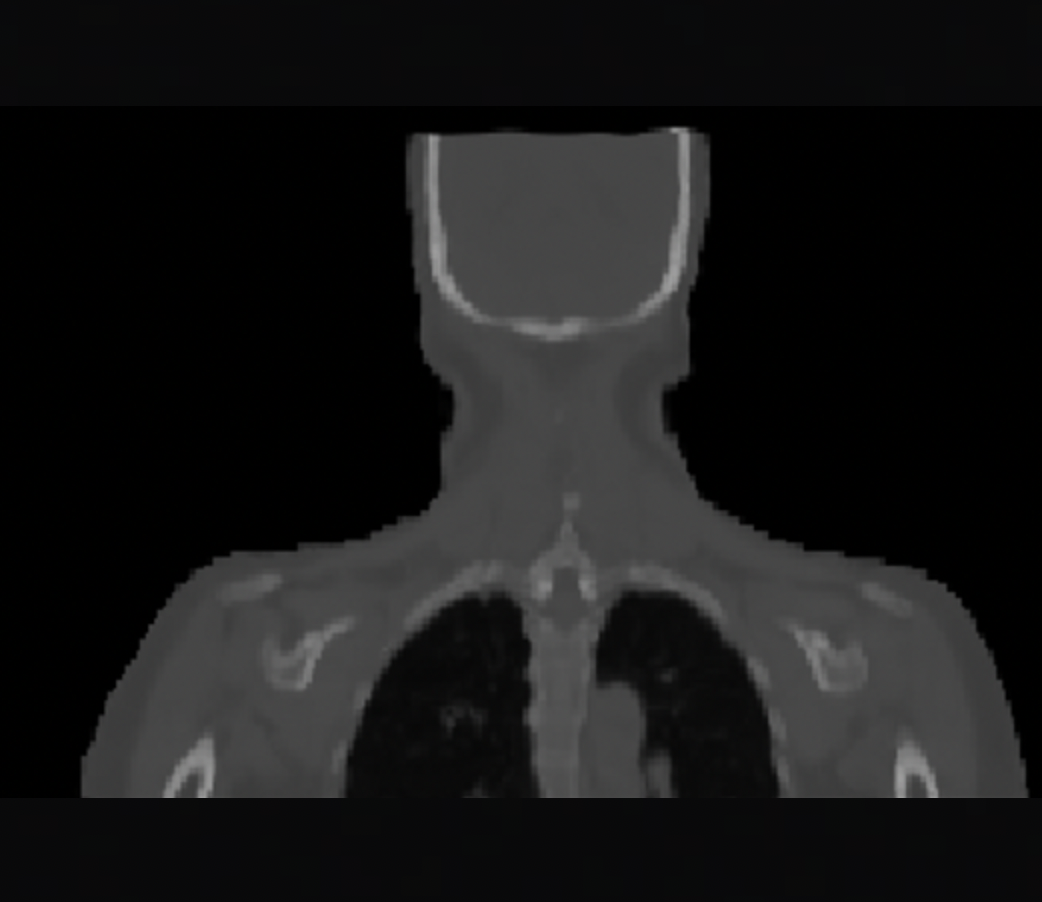} 
   
		\\
        \includegraphics[width=0.12\textwidth]{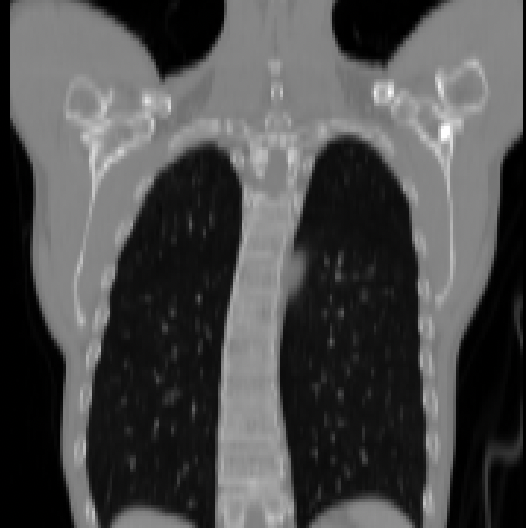}
		&\includegraphics[width=0.12\textwidth]{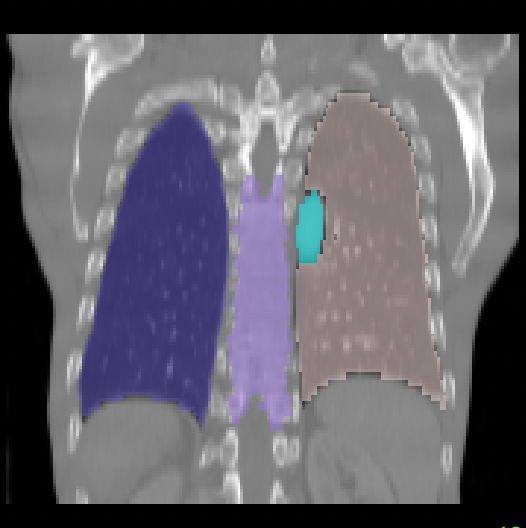}
		&\includegraphics[width=0.12\textwidth]{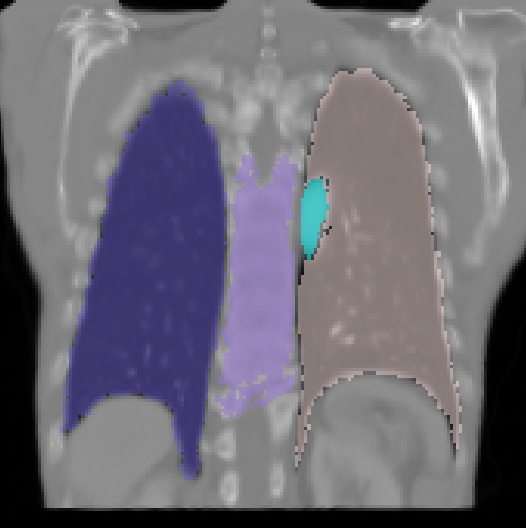}
		&\includegraphics[width=0.12\textwidth]{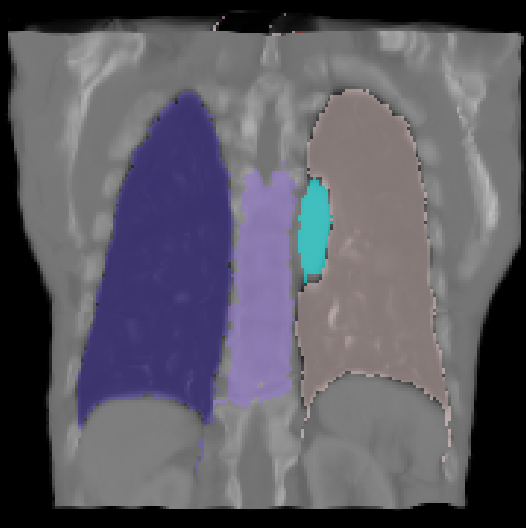}
		 \\
  
        \includegraphics[width=0.12\textwidth]{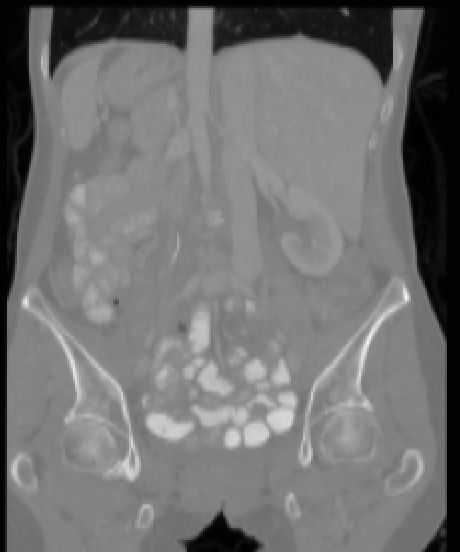}
		&\includegraphics[width=0.12\textwidth]{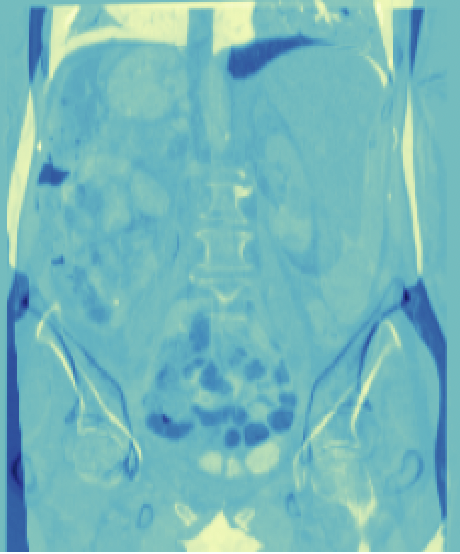}
		&\includegraphics[width=0.12\textwidth]{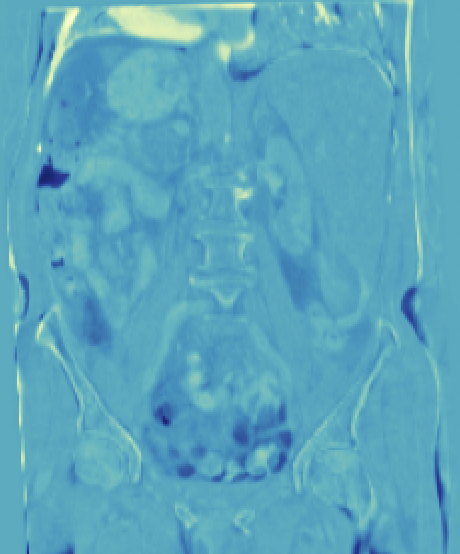}
		&\includegraphics[width=0.12\textwidth]{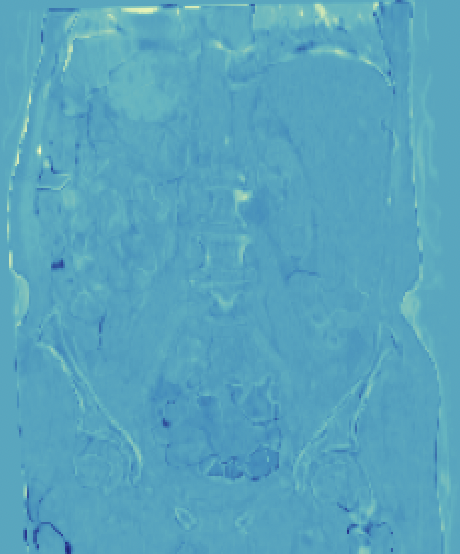} \\
  
  Moving & Fixed & Second best & SAME++\\
  
	 \end{tabular}
	\caption{\textbf{Qualitative comparisons between SAME++ and the second best methods Deeds, LapIRN and Deeds for Neck, Chest, and Abdomen registrations, respectively.} Top row: visualization of coronal head neck and warped CT slices. Middle row: Overlay of coronal chest CT (gray) and warped segmentation (color) slices.  Bottom row: Differences between warped and fixed coronal abdominal scans.}
	\label{fig:examples} 
\end{figure}

To better understand the performance of different deformable registration methods, we display organ-specific results in Fig.~\ref{fig:neck_boxplot} and Fig.~\ref{fig:chest_boxplot}. For conciseness, in the head \& neck CT dataset, we average the DICE of left and right organs into one score and calculate the median and interquartile range of DICE within each organ. In the chest CT dataset, we divide the 35 organs into 9 groups and calculate the median and interquartile range of DICE within each group. 
It is observed that SAM-deform surpasses other methods in 13 out of 17 organs or organ-groups. Some organs such as len, nerves, arteries and veins display lower DICE for all methods, this may be because they are typically small or easy to be confused with surrounding tissues. 
Qualitative examples are also shown in Fig.~\ref{fig:examples} with a clearly improved alignment of various organs after registration.

\subsection{Results of Complete SAME++ Framework}
We evaluate the performance of our complete registration method SAME++, and compare it with the complete registration pipelines (affine + deformable) of other leading registration methods, such as NiftReg, Deeds and ConvexAdam.  Note that the comparing methods here adopt their own affine transformation as SAM-affine and SAM-coarse do not exist in their methods. Table.~\ref{tab:comparsion_pipline} summarizes the quantitative results. We see that our complete SAME++ method  significantly outperforms the leading methods by improving DICE scores by 4.2\% - 8.2\%  on average over the three registration tasks. Meanwhile, it has the lowest running time with a comparable folding rate. 


\begin{table*}[htp]
\centering
\caption{Performance of the SAM instance optimization step.}
\label{tab:ablation_instance_opt}
\begin{tabular}{m{2.1cm}<{\centering} m{1.0cm}<{\centering} m{1.0cm}<{\centering} m{1.0cm}<{\centering} m{1.0cm}<{\centering} m{1.0cm}<{\centering} m{1.0cm}<{\centering}m{1.0cm}<{\centering} }
\hline
 &\multirow{2}*{Initial} & \multicolumn{3}{c}{w/o Instance Opt.} & \multicolumn{3}{c}{w/ Instance Opt.} \\ \cline{3-8} 
  ~ & ~ & DICE$\uparrow$ & $\%|J|\downarrow$ & Time(s) & DICE$\uparrow$ & $\%|J|\downarrow$ & Time(s) \\ \hline
Chest CT & 43.33 & 49.86 & 2.80 & 1.00 & 52.63 & 3.16 & 4.79 \\
Abdomen CT & 40.14 & 48.09 & 5.93 & 1.61 & 48.77 & 5.55 & 7.50 \\
HeadNeck CT & 59.99 & 64.21 & 2.92 & 0.89 & 65.58 & 2.49 &6.03  \\
\hline
\end{tabular}
\vspace{-1.em}
\end{table*}


\begin{table}[htp]
\centering
\caption{The effectiveness of stable sampling via cycle consistency (SSCC) in SAM-affine.}
\label{tab:ablation_sample_strategy}
\begin{tabular}{m{2.1cm}<{\centering} m{0.6cm}<{\centering} m{1.2cm}<{\centering} m{1.2cm}<{\centering} }
\hline
  & Initial  & w/o SSCC    & w/ SSCC \\ \hline
Chest CT  & 9.60  & 27.89 & 28.93   \\
Abdomen CT  & 25.88  & 29.67 & 29.73   \\
HeadNeck CT  & 9.21 & 31.55 & 33.56   \\
\hline
\end{tabular}
\end{table}

\subsection{Other Ablation Results}
We also conduct the following ablation studies to comprehensively understand the performance of each SAME++ component. (1) The effectiveness of SSCC in SAM-affine; (2) The importance of the regularization term in SAM-coarse; (3) The effect of using different transformations; (4) The effectiveness of SAM loss and feature; and (5) The impact of adding instance optimization. To be noted, the ablation study is conducted on the validation set of Head \& Neck CT and Chest CT datasets. Because there is no validation set in Abdomen CT dataset, we use Abdomen CT test set in the ablation study.

\textbf{Effectiveness of SSCC}: 
In SAM-affine, SSCC can help to refine the matching accuracy between corresponding points in the moving and fixed images. With SSCC, we run the iteration $5$ times to obtain a stable matching point set. As shown in Table.~\ref{tab:ablation_sample_strategy}, the stable sample strategy can improve the affine registration performance in the head \& neck and chest CT datasets by reducing the inaccurate SAM mappings. Yet, in the abdomen CT dataset, the improvement is minor. This may be because anatomic context in some abdominal organs (e.g., intestine and colon) is not unique, which results in difficulties to learn accurate SAM embedding near these organs.


\begin{table}[htp]
\centering
\caption{Performance of adding regularization in SAM-coarse}
\label{tab:ablation_sam_coarse_regularizer}
\begin{tabular}{m{2.1cm}<{\centering} m{0.6cm}<{\centering} m{0.9cm}<{\centering} m{0.9cm}<{\centering} m{0.9cm}<{\centering} m{0.9cm}<{\centering}}
\hline
 &\multirow{2}*{Initial} & \multicolumn{2}{c}{w/o Regularizer} & \multicolumn{2}{c}{w/ Regularizer} \\ \cline{3-6} 
 ~ & ~  & DICE$\uparrow$ & $\%|J|\downarrow$ & DICE$\uparrow$ & $\%|J|\downarrow$ \\ \hline
Chest CT & 28.93 & 40.41 & 22.85 & 43.33 & 0.48 \\
Abdomen CT & 29.73 & 36.29 & 35.46 & 40.14 & 2.06 \\
HeadNeck CT & 33.56 & 46.25 & 28.75 & 59.99 & 0.28 \\
\hline
\end{tabular}
\end{table}

\textbf{Regularizer in SAM-coarse}: 
We also conduct an experiment to examine the importance of adding the regularizer in SAM-coarse. As shown in Table.~\ref{tab:ablation_sam_coarse_regularizer}, adding the regularizer significantly reduces the folding percentage $\%|J|$ (on average 29.02\% to 0.94\%) in all three datasets. Moreover, it also helps to improve the registration accuracy. E.g., on average 6.84\% DICE improvement is observed. The motivation for adding a regularizer in the SAM-coarse phase is that if there exists an inaccurate match between the sampled points in the moving and fixed images, it yields a significant perturbation in the displacement field. Thus, a regularizer is required to smooth the displacement to help reducing the potentially inaccurate deformations resulting from mismatched points. 

\textbf{Transformations models}: 
In this experiment, we study how our SVF transformation model benefits the registration performance more than the Displacement Vector Field (DVF) model used in the conference version in the SAM-deform step.
The initial input in this experiment is based on the SAM-coarse results. We train two neural networks with DVF and SVF, respectively. The performance is shown in Table.~\ref{tab:ablation_transformation}. As can be seen, when SAM-deform is trained with SVF, the folding rate in the estimated map is largely reduced, leading to a more physically plausible transformation field. Meanwhile, consistent with what was observed in Table.~\ref{tab:ablation_sam_coarse_regularizer}, less folding in the transformation field also leads to more accurate registration results, i.e., a 3.43\% DICE improvement.

\textbf{Effectiveness of SAM loss and feature}: 
The ablation study for SAM loss and SAM feature in SAM-deform is shown in Table.~\ref{tab:ablation_SAMLoss}.
As shown, the best result is achieved when both the correlation feature and SAM loss are applied. We can see that the correlation feature calculated by SAM provides extra guidance for determining the deformation fields and the SAM loss provides a more semantically informed supervisory signal.

\textbf{Instance optimisation}: 
We further examine how the instance optimization step affects the final performance. Table.~\ref{tab:ablation_instance_opt} lists the registration performance with and without the SAM-based instance optimization. We see that instance optimization can consistently improve registration performance. However, as it requires a number of iterations, the running time is slightly increased.  It is a trade-off between the registration accuracy and running time, and can be used in practice according to the application requirements.

\begin{table}[!t]
\caption{Performance of SAM-deform on different transformation models.}
\label{tab:ablation_transformation}
\centering
\begin{tabular}{m{2.8cm}<{\centering}  m{1.0cm}<{\centering} m{1.0cm}<{\centering} m{1.0cm}<{\centering}}
\hline
Chest CT & Initial & DICE$\uparrow$ & $\%|J|\downarrow$ \\ \hline
w/ Displacement & 43.33 & 49.86 & 2.80 \\
w/ Diffeomorphism & 43.33 & 53.29 & 0.60 \\ \hline
\end{tabular}
\end{table}

\begin{table}[!t]
\caption{Ablation study for SAM loss and SAM feature in SAM-deform. All methods are initialized
by SAM-affine.}
\label{tab:ablation_SAMLoss}
\centering
\begin{tabular}{m{2cm}<{\centering}  m{1.6cm}<{\centering} m{2.2cm}<{\centering} m{1.0cm}<{\centering} }
\hline
Chest CT & SAM loss & SAM feature & DICE$\uparrow$ \\ \hline
Initial & \ding{55} & \ding{55} & 48.79 \\  \hline
\multirow{3}*{SAM-deform} & \ding{51} & \ding{55} & 50.43 \\ 
~ & \ding{55} & \ding{51} & 51.37 \\
~ & \ding{51} & \ding{51} & 51.99 \\ \hline
\end{tabular}
\end{table}

\section{Conclusions} \label{sec:conclusions}
In this work, we introduced the fast general SAME++ framework for medical image registration based on SAM embeddings. Specifically, we decompose image registration into four steps: affine,  coarse, deformable registration, and instance optimization, and enhance these steps by finding more coherent correspondences through the use of the SAM embeddings. 
Our SAM-affine and SAM-coarse approaches can be alternatives to optimization-based methods for registration initialization.
The SAM correlation feature and SAM loss may also be combined with any learning-based deformable registration models to serve as SAM-deform. 
We further use SAM-based instance optimization for further accuracy improvements. It can be used as a plug-and-play module for any other registration methods.
Extensive inter-patient image registration experiments using $>50$ labeled organs on chest, abdomen, and head-neck CTs datasets demonstrate the advantages of SAME++.


\bibliographystyle{IEEEtran}
\bibliography{refs}

\end{document}